\documentclass{./archive_files/amia}
\usepackage{lipsum} 
\usepackage{amsmath}
\usepackage{amssymb}
\usepackage{mathtools}
\usepackage{amsthm}
\usepackage{hyperref}
\usepackage{xcolor}
\usepackage{xspace}
\usepackage{balance}
\usepackage{multicol}
\usepackage{multirow}
\usepackage{booktabs}
\usepackage{enumitem}
\usepackage{subfig} 
\usepackage{tcolorbox}
\usepackage[utf8]{inputenc}
\tcbuselibrary{skins}

\newcommand{\header}[1]{\noindent \textbf{\textit{#1}}}
\newcommand{\modelname}{KERAP}

\begin{document}

\newcommand{\hejie}[1]{{\color{cyan} [Hejie: #1]}}

\title{KERAP: A \underline{K}nowledge-\underline{E}nhanced \underline{R}easoning \underline{A}pproach for Accurate Zero-shot Diagnosis \underline{P}rediction Using Multi-agent LLMs}

\author{Yuzhang Xie, MS$^1$, Hejie Cui, PhD$^2$, Ziyang Zhang, BS$^1$, Jiaying Lu, PhD$^1$, \\Kai Shu, PhD$^1$, Fadi Nahab, MD$^1$, Xiao Hu, PhD$^1$, Carl Yang, PhD$^1$}

\institutes{
    $^1$ Emory University, Atlanta, GA; $^2$ Stanford University, Palo Alto, CA;\\
}

\maketitle

\section*{Abstract}

\textit{Medical diagnosis prediction plays a critical role in disease detection and personalized healthcare. While machine learning (ML) models have been widely adopted for this task, 
their reliance on supervised training limits their ability to generalize to unseen cases, particularly given the high cost of acquiring large, labeled datasets.
Large language models (LLMs) have shown promise in leveraging language abilities and biomedical knowledge for diagnosis prediction. However, they often suffer from hallucinations, lack structured medical reasoning, and produce useless outputs.
To address these challenges, we propose \modelname, a knowledge graph (KG)-enhanced reasoning approach that improves LLM-based diagnosis prediction through a multi-agent architecture. Our framework consists of a linkage agent for attribute mapping, a retrieval agent for structured knowledge extraction, and a prediction agent that iteratively refines diagnosis predictions. Experimental results demonstrate that \modelname\ enhances diagnostic reliability efficiently, offering a scalable and interpretable solution for zero-shot medical diagnosis prediction.}

\section*{Introduction}
Medical diagnosis prediction, which is the task of predicting a patient’s future health risks based on their historically observed medical data such as electronic health records (EHRs), plays a vital role in enabling accurate healthcare and early interventions \cite{ghaffar2023DiagnosisPrediction, zhang2024tacco}. 
Various machine learning (ML) models, such as random forests, XGBoost, and neural networks, have been widely used for diagnosis prediction \cite{garriga2022MLforDiagnosis}. 
However, their reliance on supervised training limits their ability to make a prediction for an unseen case without an explicit training process, which is called as ``zero-shot prediction ability''. This capability is particularly crucial in scenarios where labeled medical data is scarce. Moreover, acquiring expert-labeled medical records is costly and time-intensive. Therefore, exploring zero-shot diagnosis prediction methods is essential for developing scalable and efficient predictive solutions \cite{kumar2023AIforDiagnosis}.

Recently, Large Language Models (LLMs) have shown promise in achieving zero-shot diagnosis prediction by leveraging their unprecedented language understanding abilities and rich biomedical knowledge \cite{singhal2023LLM, bhasuran2025preliminary}. By modeling patient healthcare data, LLMs can uncover hidden patterns in patient records, improving predictions for conditions such as stroke, cardiovascular diseases, and cognitive decline \cite{zhou2024LLMforDiagnosis, cui2024LLMforDiagnosis}. 
However, LLMs still face significant challenges. 
They lack structured medical reasoning capabilities, meaning they are unable to systematically process and infer relationships between medical concepts in a way that aligns with humans' clinical decision-making process.
Moreover, they are susceptible to hallucinations-generating plausible yet inaccurate predictions—which pose serious concerns in high-stakes clinical environments \cite{farquhar2024LLMChallenge}. 
Additionally, LLMs sometimes become ineffective for certain cases since they may blindly classify all cases as high-risk to minimize potential errors, leading to biased and useless predictions \cite{anibal2024LLMChallenge}. 
To ensure reliability and clinical applicability, LLMs need external guidance through structured knowledge integration. 

To overcome these challenges, knowledge graphs (KGs) have emerged as a promising solution to enhance LLM-based diagnosis prediction \cite{pan2024LLMKG}. KGs offer a structured and clinically validated framework that describes the relationships among various medical concepts, thereby enhancing the diagnostic ability of LLMs \cite{pan2023LLMKG}. By integrating KG-enhanced LLMs into diagnostic processes, the models can effectively retrieve relevant disease patterns and align their predictions with established medical knowledge \cite{gao2025LLMKG}. This symbiotic integration not only mitigates the risk of generating inaccurate or ``hallucinated'' information but also enhances the interpretability and reliability of the diagnostic outputs \cite{jia2025LLMKG}.  
However, existing KG-enhanced diagnosis prediction approaches typically rely on directly prompting LLMs with KG knowledge and EHR data, whose performance could be improved by incorporating a more complex medical reasoning process \cite{gao2025LLMKG, jia2025LLMKG, wu2024LLMKG}. For instance, a multi-agent framework could decompose complex medical reasoning into specialized tasks, allowing different agents to focus on different tasks such as linking medical attributes across different data sources, retrieving knowledge, and predicting a patient's diagnosis. This reduces cognitive overload for a single model and allows for more structured, task-specific optimizations. Moreover, a multi-stage conversational reasoning framework could enhance robustness by enabling iterative refinement, allowing the LLM to focus on different parts of the extracted knowledge and reflect on its prediction.

In this work, we propose \modelname, a KG-enhanced multi-stage reasoning approach for accurate zero-shot diagnosis prediction using LLMs. \modelname\ operates through a multi-agent architecture: a linkage agent, a retrieval agent, and a prediction agent. The linkage agent initiates the process by establishing connections between medical attributes—such as symptoms, diagnoses, and treatments—extracted from real-world patient datasets (EHRs) and a comprehensive biomedical KG. 
LLMs are considered state-of-the-art methods to link different medical attributes \cite{xie2024promptlink}. By mapping raw patient data to structured medical knowledge accurately, this agent ensures that the retrieved information is contextually relevant and aligned with clinical reasoning. 
Subsequently, the retrieval agent queries the KG to extract and summarize attribute-related knowledge, categorizing them into two distinct perspectives: positive knowledge (e.g., ``symptom X indicates condition Y'') for inclusion criteria, and negative knowledge (e.g., ``symptom X rules out condition Z'') for exclusion criteria. 
This step reduces LLM hallucinations and enhances reasoning consistency by ensuring the LLM is guided by structured, validated knowledge rather than unverified text. These related knowledge are then formatted as contextual prompts to provide structured knowledge and reduce hallucinations.
Finally, the prediction agent integrates patient-specific data with the retrieved external knowledge, engaging in multi-stage conversation steps to refine its understanding and produce a robust, patient-tailored diagnosis prediction. 
In our experiments, \modelname\ demonstrates superior performance, achieving higher accuracy and reliability in zero-shot scenarios. Our approach contributes to the advancement of predictive modeling in clinical artificial intelligence, offering a scalable solution that supports early disease detection, personalized risk assessment, and improved patient outcomes.

\section*{Related Work}
\header{LLM Reasoning.} 
Large Language Models (LLMs), including OpenAI’s GPT and Meta's LLaMa, have transformed various domains by enhancing reasoning capabilities \cite{zhou2024LLMforDiagnosis}. To address complex tasks such as diagnosis prediction, diverse reasoning strategies can be integrated with LLMs to support medical reasoning. For example, Direct Prompting serves as a basic approach, eliciting immediate responses from LLMs. More advanced Step-by-Step techniques, such as Chain-of-Thought \cite{wei2022cot}, decompose problems into logical intermediate steps. Additionally, KG-Augmented Prompting further incorporates external, structured knowledge to improve factual accuracy and interoperability \cite{gao2025LLMKG}. Iterative Prompting advances by allowing predictions to be refined over multiple stages through reflective evaluation \cite{renze2024iterative}.
These strategies collectively enable LLMs to generate outputs that are coherent, contextually relevant, and clinically accurate.

\header{LLM for Diagnosis Prediction.}
LLMs have demonstrated substantial potential in enhancing medical diagnosis prediction by analyzing extensive clinical data and patient records. 
Researchers have explored direct prompting LLMs for diagnosis prediction \cite{chiu2024LLMDiagnosis, sonoda2024LLMDiagnosis}. 
Additionally, some approaches leverage pre-trained medical-specialized LLMs, such as HuatuoGPT\cite{chen2023huatuogpt}.
Some approaches combine the LLM with other trained classification ML models (e.g., Logistic Regression, XGBoost) \cite{xu2024LLMDiagnosis, wang2024LLMDiagnosis, jin2024LLMDiagnosis, zhu2024LLMDiagnosis}.
Another promising direction is in-context learning, where LLMs adapt to medical tasks by utilizing few-shot examples, as seen in CPLLM \cite{ben2024CPLLM} and EHR-CoAgent \cite{cui2024LLMforDiagnosis}.
Furthermore, some methods integrate biomedical KGs with LLMs, such as Dr.Knows \cite{gao2025LLMKG}, medIKAL \cite{jia2025LLMKG}, and ICP \cite{wu2024LLMKG}, improving interpretability and diagnostic precision by offering structured explainable relations.
However, these approaches often struggle with adaptability to unseen cases, as they either depend on simple direct prompting, require extensive labeled data for fine-tuning or training, or lack structured multi-step reasoning to iteratively refine predictions.
\section*{Method}

\header{Task Definition.}
We aim to predict whether a patient will be diagnosed with a specific disease at their next clinical visit, based on medical attributes recorded during the current visit. Formally, let $i$ denote the $i$-th clinical visit of a patient, and let $\mathbf{r}_i$ be the set of EHR entries associated with this visit: $\mathbf{r}_i = \{ a_{i,0}, a_{i,1}, \dots, a_{i,j} \}$, where each $a_{i,j}$ represents a medical attribute such as a diagnosis (e.g., heart failure) or medication (e.g., desmopressin). These attributes capture crucial aspects of a patient's diagnosis and treatment during the visit. Given $\mathbf{r}_i$, the goal is to predict whether a target disease $d$ will be diagnosed at the next visit.

\begin{figure}
    \centering
    \includegraphics[width=1\linewidth]{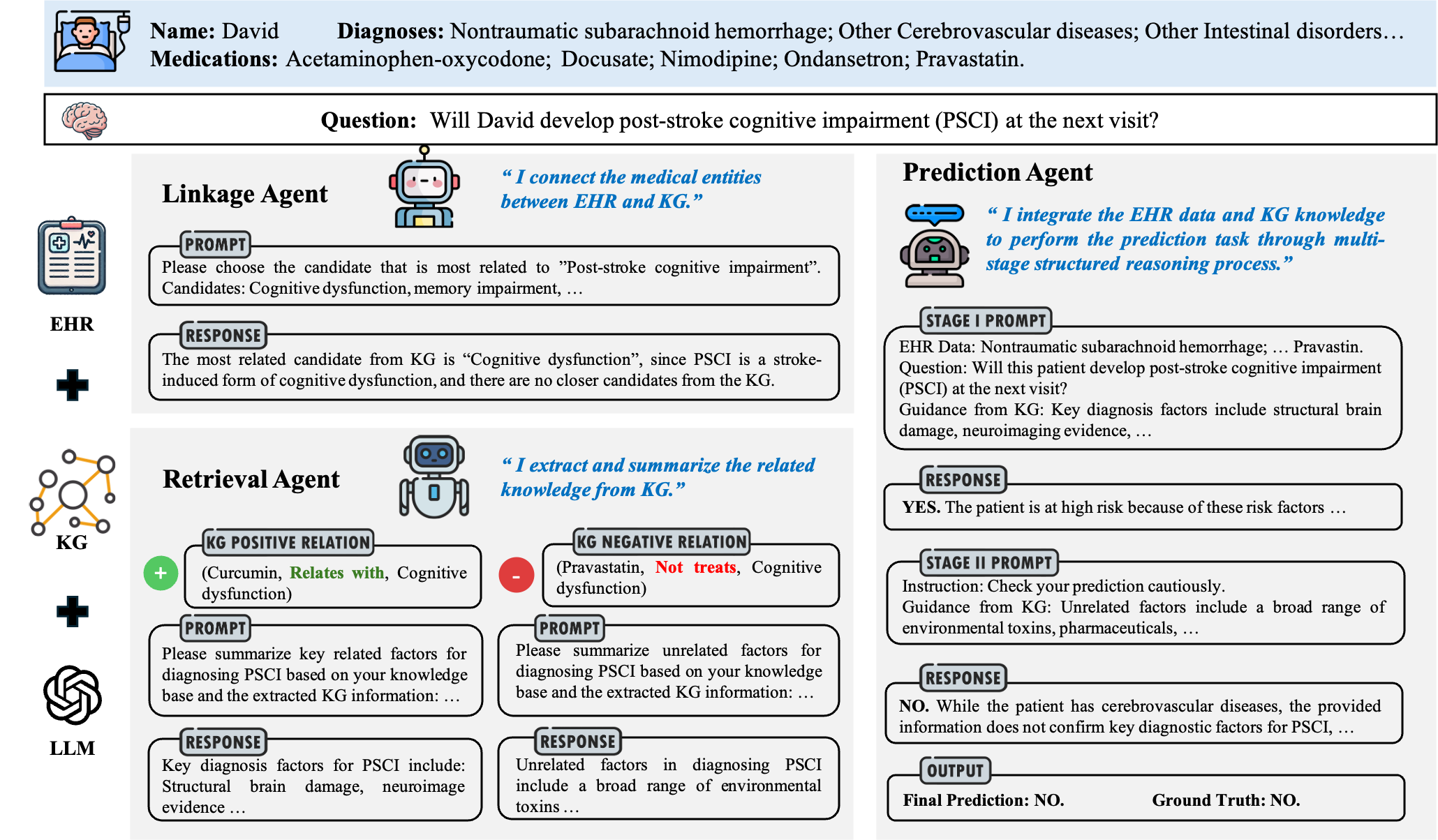}
    \vspace{-2em}
    \caption{Framework of \modelname.}
    \label{fig:framework}
\end{figure}

\header{Framework.}
The \modelname\ framework is illustrated in Fig.\ref{fig:framework}. 
It adopts a multi-agent architecture, with GPT-4o-mini supporting each agent. The framework comprises three key components: the linkage agent, the retrieval agent, and the prediction agent.
The linkage agent $A_{LK}$ locates the predicted disease in a biomedical KG, serving as a prerequisite for the subsequent retrieval agent to access relevant biomedical knowledge for that disease. For example, the disease expression ``Post-stroke cognitive impairment'' in the question text is linked to the most appropriate KG entity ``Cognitive dysfunction'' by $A_{LK}$,  since there are no closer candidates from the KG. The retrieval agent $A_{RT}$ queries the KG to extract and summarize relevant information, categorizing the results into positive knowledge relations, e.g., \textit{(Curcumin, Relates with, Cognitive dysfunction)}, and negative relations, e.g., \textit{(Pravastatin, Not treats, Cognitive dysfunction)}, to support inclusion and exclusion criteria. Finally, the prediction agent $A_{PR}$ integrates patient records with the extracted structured knowledge and employs multi-stage reasoning to perform zero-shot diagnosis prediction, producing a clear ``YES'' or  ``NO'' outcome.

\header{Linkage Agent.}
To connect the predicted disease entity $d$ from EHR with the relevant KG entity $x$, we leverage a linkage agent $A_{LK}$ inspired by PromptLink \cite{xie2024promptlink}.  
To design effective and cost-efficient prompts within the constraints of context length, linking candidates are first generated before the linkage agent performs entity disambiguation. Specifically, we utilize a pre-trained language model, SAPBERT \cite{liu2020sapbert}, to generate embeddings $Z_{lm, d} \in \mathbb{R}^{d_{lm}}$ for the EHR entity $d$ and $Z_{lm, x} \in \mathbb{R}^{d_{lm}}$ for the KG entity $x$ ($d_{lm}=768$ from SAPBERT).  
Then we compute the cosine similarity $SIM \in [0, 1]$ between $Z_{lm, d}$ and $Z_{lm, x}$, selecting the top-$LC$ candidates (where $LC = 10$) with the highest similarity scores as linking candidates. These candidates serve as input for the linkage agent, which further generates the entity mapping. 
By analyzing the semantic meaning and contextual relevance of the candidates, the linkage agent chooses the KG entity $x$ as the most appropriate one to link for the EHR medical entity $d$.  
This agent $A_{LK}$ utilizes both deep embeddings and LLM-based reasoning to ensure efficiency and precision in linking medical entities. 
By ensuring accurate disease-to-KG mapping, \modelname\ enables accurate integration of biomedical knowledge from KG.
The process is formally defined as:  
\begin{equation}
\label{equ:promptlink}
    x = A_{LK}(d).
\end{equation}

\header{Retrieval Agent.}
After linking a disease entity $d$ from the EHR to its corresponding KG entity $x$ by $A_{LK}$, relevant knowledge is extracted from the KG by the retrieval agent $A_{RT}$. The KG is a multi-relational graph denoted as $\text{KG} = (X, R, RT)$, where $X$ represents the set of entities (nodes), $R$ denotes the set of relations (edges), and $RT$ contains the relational triples.  
The relations of the linked KG entity $x$ are categorized into positive knowledge relations $RT_{x, +}$ and negative knowledge relations $RT_{x, -}$ based on their semantic meaning. 
For example, \textit{(Curcumin, Relates with, Cognitive dysfunction)} is classified as positive knowledge relation, indicating that curcumin relates with cognitive dysfunction. Conversely, \textit{(Pravastatin, Not treats, Cognitive dysfunction)} is a negative knowledge relation, signifying that pravastatin does not treat cognitive dysfunction.
Next, the retrieval agent $A_{RT}$ is prompted to summarize the extracted relations $RT_{x, +}$ and $RT_{x, -}$. These summarized relations encapsulate relevant knowledge for inclusion or exclusion criteria. The extracted knowledge for $d$ is represented as $\mathcal{KN}_{d}$ in:  
\begin{equation}
\label{equ:retrieval}
   \mathcal{KN}_{d} = \{\mathcal{KN}_{d,+}; \mathcal{KN}_{d,-}\} = A_{RT}(RT_{x}) = A_{RT}(\{RT_{x, +}; RT_{x, -}\}).
\end{equation}

\header{Prediction Agent.}
After retrieving the relevant knowledge $\mathcal{KN}_{d}$ for disease $d$, it is integrated with a patient-specific record $r_i$ as prompt for the prediction agent $A_{PR}$. The prediction agent is then tasked with determining whether the patient will develop disease $d$ at the next visit, completing the diagnosis prediction task.  
$A_{PR}$ follows a multi-stage reasoning process. In Stage I, it considers the patient's visit data $r_i$ along with positive relations $\mathcal{KN}_{d, +}$ to make an initial prediction, leveraging both patient-specific visit data and enlightening inclusion criteria from external knowledge. However, since LLMs sometimes become ineffective and blindly classify all cases as high-risk, $A_{PR}$ refines its prediction in the second stage. In Stage II, $A_{PR}$ combines the Stage I prompt and output with the negative relations $\mathcal{KN}_{d, -}$, leveraging self-reflection and exclusion criteria from external knowledge to eliminate misclassified cases. 
The prediction process is formulated as follows, where $\mathcal{PR}_{i,d}$ represents the prediction result for patient visit $i$ and disease $d$:  
\begin{equation}
\label{equ:prediction}
   \mathcal{PR}_{i, d} = A_{PR}(\{r_i;\ \mathcal{KN}_{d}\}) = \text{``YES'' or ``NO''}.
\end{equation}

\section*{Experiments}

\header{EHR Datasets.}
In our experiments, we perform diagnosis prediction tasks for four datasets: Pneumonia (PNA), Chronic Kidney Disease (CKD), Congestive Heart Failure Nonhypertensive (CHF), and Post-Stroke Cognitive Impairment (PSCI). Patient-specific EHR data for Pneumonia are sourced from MIMIC-III \cite{johnson2016mimic}, while PSCI data come from PROMOTE. 
MIMIC-III is a public dataset containing over 53,000 patient records from critical care units at Beth Israel Deaconess Medical Center between 2001 and 2012. Following the settings of Xu et al. \cite{xu2023hypergraph}, we select 12,353 patient encounters, incorporating 7,408 key medical attributes, including 845 diagnoses, 4,522 prescriptions, 2,021 procedures, and 20 services. 
PROMOTE is a private dataset containing records of stroke patients treated from 2012 to 2021 \cite{promote}. We extract data on 7,780 patients and 2,595 clinical attributes, including 1,480 ICD-10 diagnosis codes and 1,115 prescribed medications recorded up to each patient’s discharge following their index stroke.

\header{KG Data Source.}
We use a large-scale public KG, iBKH \cite{su2023ibkh}, as the primary KG dataset. iBKH integrates data from various biomedical KGs, offering a comprehensive resource with 2,384,501 entities across 11 categories, including drugs, diseases, symptoms, genes, and pathways, \textit{etc}. Moreover, with over 48 million relation triples, iBKH facilitates deeper insights into complex biological interactions.

\header{Baselines.}
We evaluate the performance of \modelname\ against the following baseline methods:
\begin{itemize}[topsep=0pt, parsep=0pt, itemsep=0pt]
    \item Direct Prompting: Directly prompting LLM with patient's EHR data.
    \item Step-by-Step Prompting: Prompting the LLM with a patient's EHR data and prediction steps in a single stage using a representative Step-by-Step reasoning technique, Chain-of-Thought \cite{wei2022cot}.
    \item Pre-trained Medical LLM: Prompting patient's EHR data by using a representative, open-source pre-trained medical LLM, HuatuoGPT2-7B \cite{chen2023huatuogpt}. 
    \item KG-Augmented Prompting: Prompting LLM with patient's EHR data and relevant KG knowledge,  following methodologies similar to Dr. Knows \cite{gao2025LLMKG}, medIKAL \cite{jia2025LLMKG}, and ICP \cite{wu2024LLMKG}.
     \item Iterative Prompting: Prompting LLM with patient's EHR data and multi-stage reasoning,  following methodologies similar to \modelname, but with no information from KG.
\end{itemize}

\header{Implementation Details.}
We implement our empirical study in Python, leveraging libraries such as Hugging Face, OpenAI, and Scikit-learn. For the zero-shot diagnosis experiments, we primarily use the Azure OpenAI Service with the GPT-4o-mini model, except for the baseline method "Pre-trained Medical LLM", which utilizes the specialized medical model HuatuoGPT2-7B. All evaluated methods are not specifically trained with our four curated datasets.
Performance is evaluated using two standard classification metrics: Accuracy (ACC) and F1-score. Considering dataset imbalance, the F1-score is computed as a weighted average across both classes. Each experiment is repeated five times to ensure reliability, with the mean and standard deviation reported. By utilizing Azure OpenAI, we ensure secure processing of sensitive patient data, adhering to regulatory standards such as HIPAA and GDPR.
\footnote{The code, retrieved knowledge, and detailed prompts of this paper are available at \href{https://github.com/constantjxyz/KERAP}{https://github.com/constantjxyz/KERAP}.}

\header{Zero-shot Diagnosis Results.}

\vspace{2pt}
\begin{table}[htbp]
    \centering   
    \caption{Zero-shot Performance (\%) of LLM-based diagnosis prediction methods (mean±std across five runs).}
    \label{tab:results}
    \vspace{-5pt}
    \resizebox{1.0\linewidth}{!}{
    \begin{tabular}{ccccccccc}
        \hline
        \multirow{3}{*}{Method} & \multicolumn{2}{c}{PSCI} & \multicolumn{2}{c}{PNA} & \multicolumn{2}{c}{CKD} & \multicolumn{2}{c}{CHF} \\
        &\multicolumn{2}{c}{(Prevalence:22.30\%)} & \multicolumn{2}{c}{(Prevalence:21.71\%)} & \multicolumn{2}{c}{(Prevalence:25.47\%)} &\multicolumn{2}{c}{(Prevalence:39.20\%)} \\
        \cmidrule(lr){2-3} \cmidrule(lr){4-5}\cmidrule(lr){6-7}\cmidrule(lr){8-9}
        & ACC & F1 & ACC & F1 & ACC & F1 & ACC & F1 \\
        \hline
        Direct Prompting & 21.31$\pm$0.58 & 7.78$\pm$0.30 & 33.77$\pm$2.11 & 32.94$\pm$2.50 & 33.46$\pm$0.55 & 25.41$\pm$1.02 & 46.15$\pm$0.92 & 36.32$\pm$1.79 \\
        Step-by-Step Prompting & 22.22$\pm$0.62 & 7.89$\pm$0.41 & 32.59$\pm$1.24 & 30.56$\pm$1.64 & 32.70$\pm$0.87 & 24.21$\pm$1.68 & 43.55$\pm$1.63 & 37.51$\pm$1.77 \\
        Pre-trained Medical LLM & 28.90$\pm$0.56 & 25.98$\pm$0.66 & 27.43$\pm$0.90 & 22.16$\pm$1.27 & 39.55$\pm$1.16 & 37.07$\pm$1.48 & 41.32$\pm$0.31 & 33.06$\pm$0.82 \\
        KG-Augmented Prompting & 27.78$\pm$0.56 & 23.98$\pm$0.82 & 70.88$\pm$0.30 & 70.54$\pm$0.42 & 47.01$\pm$0.62 & 46.20$\pm$0.81 & 47.02$\pm$0.61 & 38.20$\pm$1.01 \\
        Iterative Prompting & 24.01$\pm$1.39 & 15.19$\pm$0.97 & 66.61$\pm$2.69 & 67.77$\pm$1.46 & 70.60$\pm$2.20 & 72.31$\pm$2.05 & 58.45$\pm$2.62 & 57.67$\pm$2.30 \\
        \modelname\ (Ours) & \textbf{72.44$\pm$0.71} & \textbf{68.98$\pm$0.74} & \textbf{74.24$\pm$0.39} & \textbf{71.49$\pm$0.18} & \textbf{76.16$\pm$0.45} & \textbf{77.42$\pm$0.43} & \textbf{70.80$\pm$1.53} & \textbf{71.06$\pm$1.37} \\
        \hline
    \end{tabular}
    }
\end{table}

The zero-shot diagnosis prediction results of different LLM-based methods across our four EHR datasets and five runs are shown in Table \ref{tab:results}. Baseline methods, including Direct Prompting, Step-by-Step Prompting, and Pre-trained Medical LLM, perform poorly, with Accuracy and F1-score all below 50\%. A key reason is that LLMs are often ineffective for these diseases, blindly misclassifying most or even all cases as high-risk. This bias is especially problematic given that disease prevalence in the datasets is below 40\%, rendering such predictions impractical and clinically uninformative. Moreover, these methods lack structured medical reasoning, preventing them from systematically processing and inferring relationships between medical concepts in alignment with clinical decision-making. They also generate hallucinations, further degrading their performance.

In contrast, methods incorporating KG knowledge (KG-Augmented Prompting) and multi-stage reasoning (Iterative Prompting) achieve better results, particularly on CKD and PNA datasets, indicating that structured knowledge and more advanced reasoning help mitigate the limitations of simpler baselines. Finally, our proposed model, \modelname, outperforms all methods across all datasets, demonstrating its effectiveness in zero-shot diagnosis. This improvement can be attributed to two key design choices: leveraging KG knowledge and enhancing reasoning through a multi-stage process, which reduce hallucinations and encourage LLMs to reason more like human clinicians. Additionally, with guidance from KG-extracted positive and negative knowledge, \modelname\ produces more accurate and reliable diagnoses.

\header{Cost and Efficiency.}
\vspace{5pt}
\begin{table}[htbp]
    \centering
    \caption{Comparison of cost and efficiency of different LLM-based diagnosis prediction methods.}
    \label{tab:cost_comparison}
    \vspace{-5pt}
    \resizebox{0.9\linewidth}{!}{
    \begin{tabular}{cccccc}
        \hline
        Method & Total Tokens (million) & Token Cost (\$) & Time Cost (h) & ACC Performance (\%) & F1 Performance (\%) \\
        \hline
        Direct Prompting & 19.95 & 3.19 & 8.88 & 33.67 & 25.61 \\
        Step-by-Step Prompting & 20.49 & 3.35 & 9.61 & 32.77 & 25.04 \\
        Pre-trained Medical LLM & 21.03 & 0.00 & 0.69 & 34.30 & 29.56 \\
        KG-Augmented Prompting & 39.17 & 6.07 & 9.03 & 48.17 & 44.73 \\
        Iterative Prompting & 42.22 & 6.73 & 17.74 & 54.92 & 53.24 \\
        \modelname\ (Ours) & 72.05 & 11.20 & 17.59 & 73.41 & 72.24 \\
        \hline
    \end{tabular}
    }
\end{table}

To evaluate the cost and efficiency of \modelname, we summarize the token cost, time cost, and performance of various methods across four EHR datasets in Table~\ref{tab:cost_comparison}. Reported costs represent the total across all datasets, averaged over five runs. Accuracy and F1 scores are averaged across the four datasets.
Among the methods, Pre-trained Medical LLM incurs the lowest token and time costs, as it is open-source and runs inference locally. For other methods, token costs are estimated based on GPT-4o-mini pricing \footnote{The API price is from \href{https://platform.openai.com/docs/pricing}{https://platform.openai.com/docs/pricing}}. Among the GPT-based approaches, Direct Prompting and Step-by-Step Prompting require the fewest tokens and have the lowest costs. In contrast, KG-Augmented Prompting and Iterative Prompting involve more complex reasoning, leading to higher token usage and longer computation times. Our proposed method, \modelname, exhibits the highest token consumption (72.05 million), token cost (\$11.20), and time cost (17.59h), reflecting its more complex reasoning. However, given the large patient population across the four datasets (over 44,000 in total) and its strong performance, \modelname\ remains a practical and valuable approach for real-world applications.

\header{Impact of LLM Choice.}

\begin{figure}[htbp]
    \centering
    \includegraphics[width=1.0\linewidth]{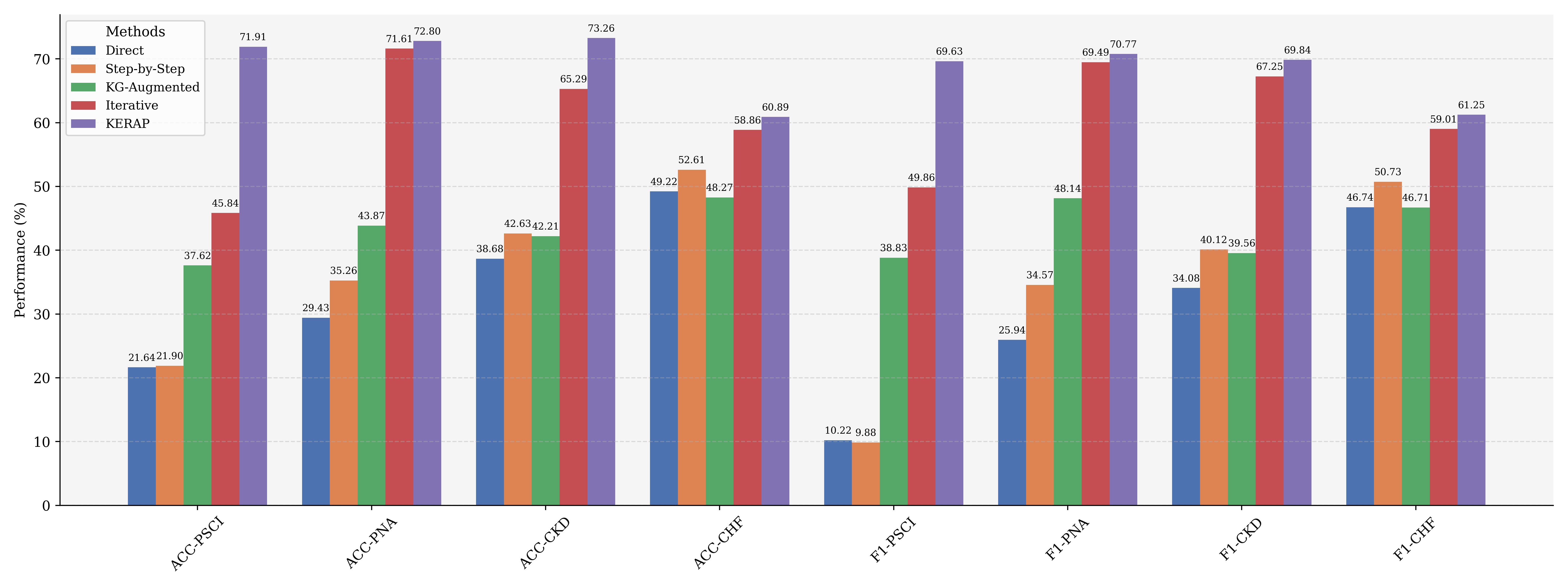}
    \vspace{-2em}
    \caption{Zero-shot performance (\%) of different LLM-based diagnosis prediction methods using \textit{GPT-3.5-turbo}. The blue, orange, green, red, and purple bars represent the results of five compared methods: Direct Prompting, Step-by-Step Prompting, KG-Augmented Prompting, Iterative Prompting, and \modelname, respectively.}
    \label{fig:gpt35}
\end{figure}

\begin{figure}[htbp]
    \centering
    \includegraphics[width=1.0\linewidth]{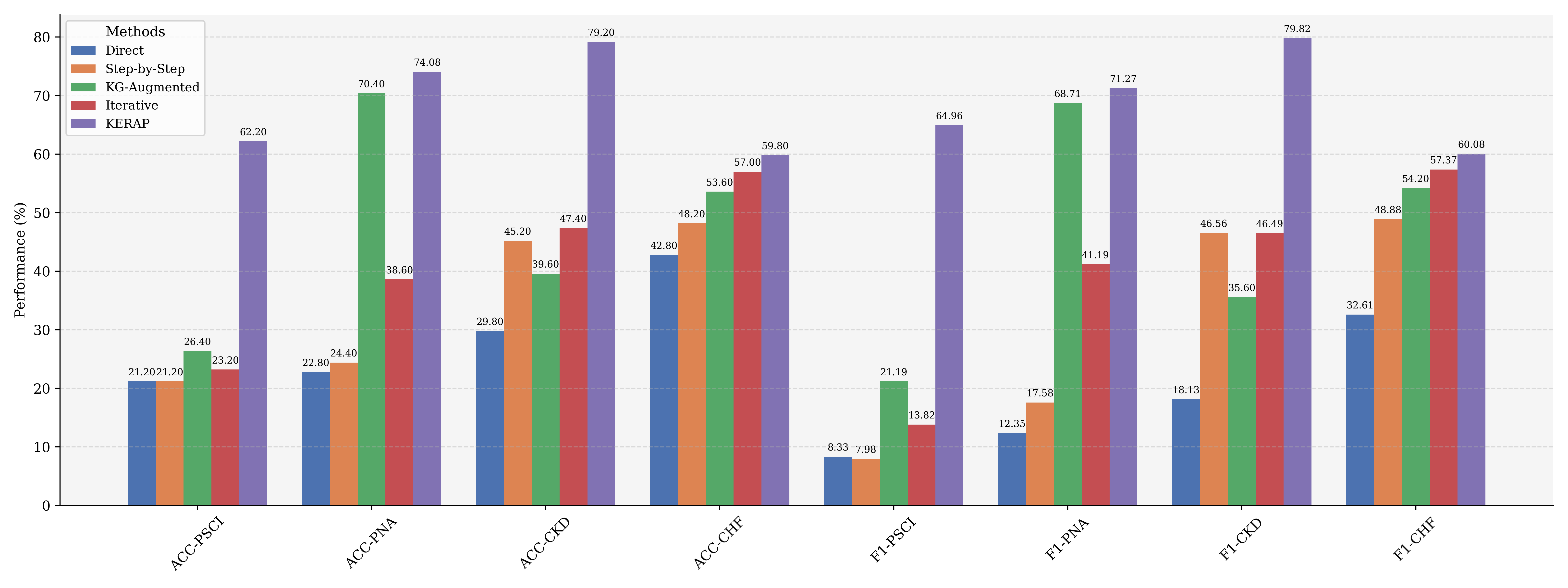}
    \vspace{-2em}
    \caption{Zero-shot performance (\%) of different LLM-based diagnosis prediction methods using \textit{GPT-4o}. The blue, orange, green, red, and purple bars represent the results of five compared methods: Direct Prompting, Step-by-Step Prompting, KG-Augmented Prompting, Iterative Prompting, and \modelname, respectively.}
    \label{fig:gpt4o}
\end{figure}

To further evaluate the performance of different diagnosis prediction methods, we conduct experiments using various LLMs, replacing GPT-4o-mini with GPT-3.5-turbo and GPT-4o. 
Taking all four datasets' performance into consideration, as shown in Figure~\ref{fig:gpt35} (GPT-3.5-turbo), Figure~\ref{fig:gpt4o} (GPT-4o), and Table~\ref{tab:results} (GPT-4o-mini), we observe that \modelname\ consistently outperforms all other methods in both Accuracy and F1-score on all four datasets, demonstrating superior predictive capability and robustness across different LLMs. Additionally, GPT-3.5-turbo and GPT-4o generally underperform compared to GPT-4o-mini, reinforcing our decision to adopt GPT-4o-mini as the primary model for zero-shot diagnosis. The performance gap is likely due to GPT-3.5-turbo’s limited reasoning ability and knowledge base, while GPT-4o classifies an excessive number of cases as high-risk in certain datasets (e.g., PSCI and CHF), resulting in biased predictions.

Moreover, as shown in Table~\ref{tab:cost_gpt35_gpt4o} (GPT-3.5-turbo and GPT-4o) and Table~\ref{tab:cost_comparison} (GPT-4o-mini), both GPT-3.5-turbo and GPT-4o incur significantly higher monetary and time costs compared to GPT-4o-mini \footnote{Pricing from \href{https://platform.openai.com/docs/pricing}{https://platform.openai.com/docs/pricing}}. Considering the trade-off between performance, cost, and efficiency, GPT-4o-mini stands out as the most practical and effective choice for zero-shot diagnosis prediction on our datasets, further validating our selection.

\begin{table}[htbp]
    \centering
    \vspace{10pt}
    \caption{Comparison of cost and efficiency of different LLM-based diagnosis prediction methods.}
    \label{tab:cost_gpt35_gpt4o}
    \vspace{-5pt}
    \resizebox{1.0\linewidth}{!}{
    \begin{tabular}{ccccccc}
        \hline
        \multirow{2}{*}{Method} & \multicolumn{3}{c}{GPT-3.5-turbo} & \multicolumn{3}{c}{GPT-4o} \\
        \cmidrule(lr){2-4} \cmidrule(lr){5-7} 
        & Total Tokens (million) & Token Cost (\$) & Time Cost (h) & Total Tokens (million) & Token Cost (\$) & Time Cost (h) \\
        \hline
        Direct Prompting & 19.34 & 7.38 & 8.40 & 19.55 & 51.74 & 10.42 \\
        Step-by-Step Prompting & 23.53 & 12.17 & 9.76 & 20.42 & 54.77 & 12.11 \\
        KG-Augmented Prompting& 42.22 & 20.00 & 11.25 & 39.25 & 101.57 & 12.05 \\
        Iterative Prompting & 48.40 & 25.19 & 18.21 & 42.30 & 112.79 & 21.67 \\
        \modelname\ (Ours) & 82.77 & 42.43 & 20.98 & 75.11 & 196.48 & 23.64 \\
        \hline
    \end{tabular}
    }
\end{table}

\newpage
\header{Comparison with Supervised Learning Methods.}

\begin{figure}[htbp]
    \centering
    \subfloat[ACC]{%
        \includegraphics[width=0.46\textwidth]{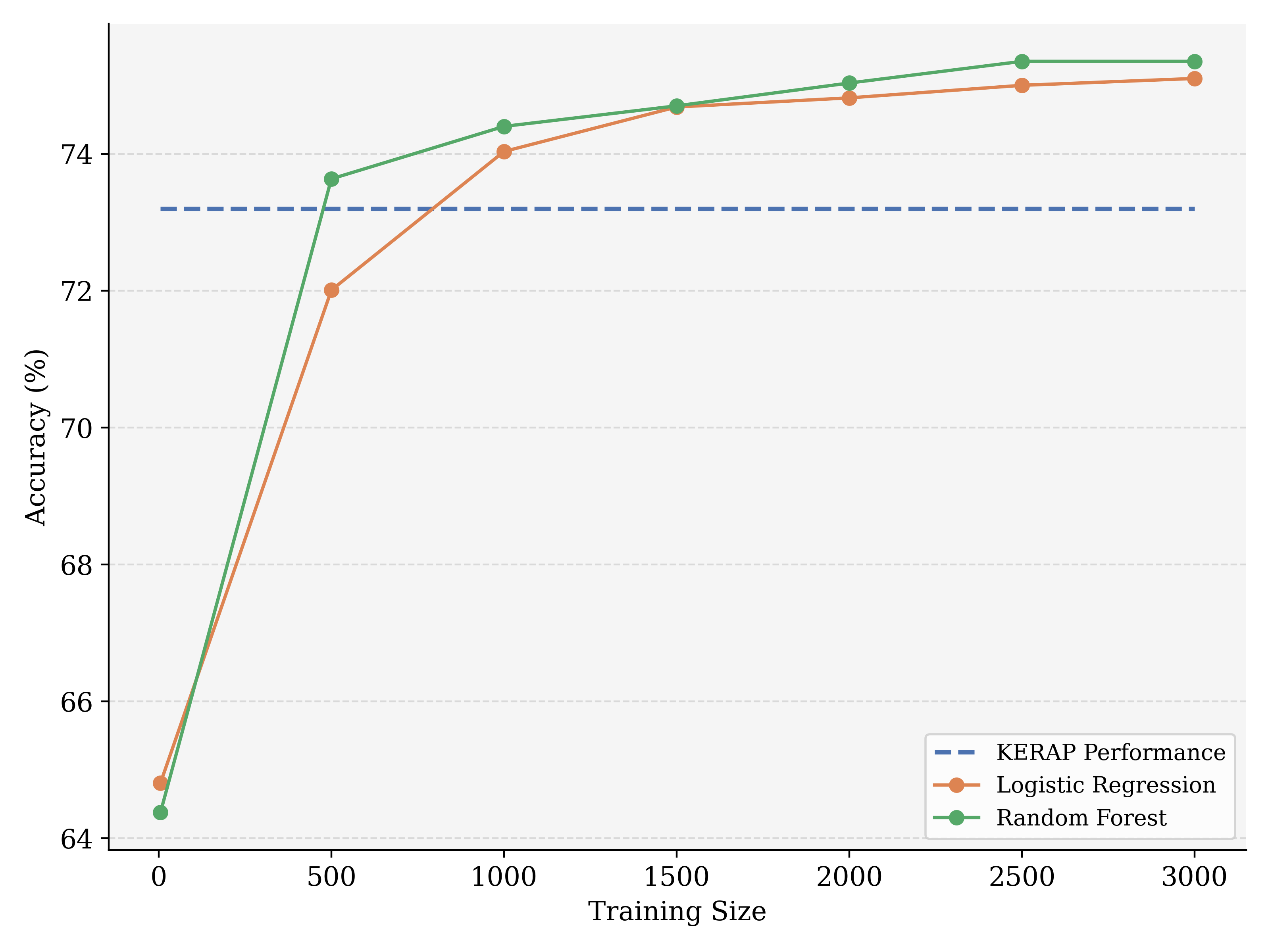}
    }
    \hspace{0.05\textwidth} 
    \subfloat[F1]{%
        \includegraphics[width=0.46\textwidth]{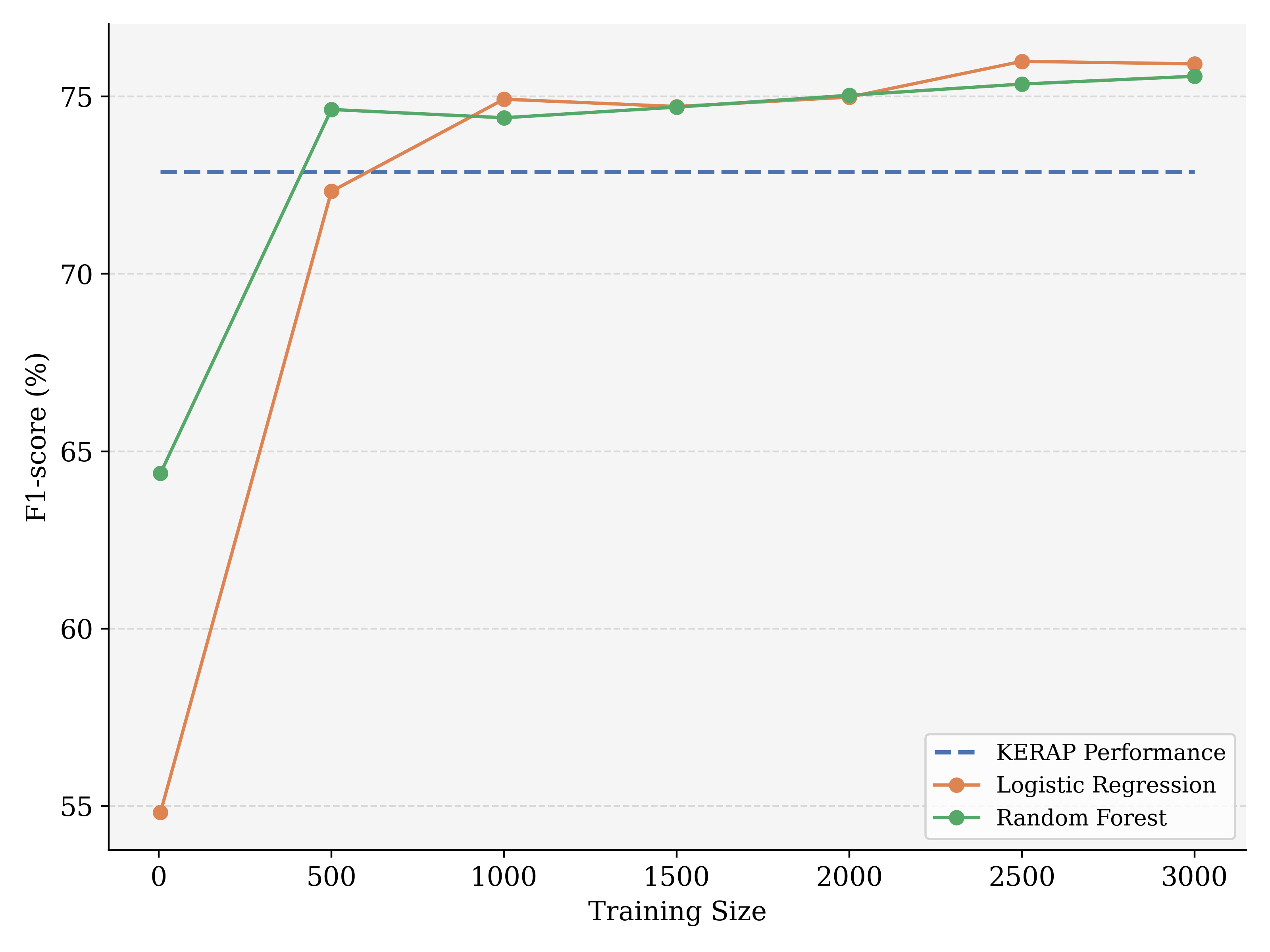}
    }
    \vspace{-10pt}
    \caption{Performance comparison of \modelname\ with supervised learning methods in terms of Accuracy (a) and F1-score (b). In the figures, the blue horizontal dashed line indicates the diagnosis prediction performance of \modelname, which operates without any training samples as \textit{zero-shot} prediction. The green dotted line represents the performance of random forest across different training sample sizes, while the orange dotted line represents logistic regression under varying training samples. }
    \label{fig:supervised}
\end{figure}

To further assess the performance of \modelname, we compare it with two representative supervised learning methods: logistic regression and random forest. We evaluate these models across varying training sample sizes to measure their effectiveness relative to \modelname. For a fair comparison, we hold out a fixed test set of 500 samples for all methods and report the average Accuracy and F1-score across four datasets. The results, shown in Figure \ref{fig:supervised}, indicate that logistic regression and random forest achieve slightly higher performance than \modelname\ (73.20\% in ACC, 72.88\% in F1), surpassing it by approximately 2\%. However, this improvement comes at the cost of requiring a substantial amount of high-quality labeled data—typically 500–1000 samples. Obtaining expert-labeled medical records is both costly and time-consuming. In contrast, \modelname\ operates as a zero-shot diagnosis prediction method, eliminating the need for labeled training data while achieving performance comparable to supervised models trained on large labeled datasets. This makes \modelname\ a scalable and efficient solution, particularly for diseases where labeled data is scarce.

\newpage
\header{Case Studies.}

To demonstrate the effectiveness of \modelname, we present two diagnosis prediction cases - Case A and Case B. Each case includes the prompt, prediction results, and reasoning process generated by GPT-4o-mini. We analyze the cases and advantages of \modelname\ from the perspectives described below.

\begin{itemize}[topsep=0pt, itemsep=0pt, parsep=0pt, leftmargin=6pt]

    \item \textbf{Zero-shot capability:} Unlike supervised methods, \modelname\ requires no training or fine-tuning. It directly utilizes patient data to make predictions for unseen cases, as demonstrated in our examples.

    \item \textbf{Factuality guided by KG:} \modelname\ enhances factuality through KG-grounded reasoning:
    \begin{itemize}[topsep=0pt, itemsep=0pt, parsep=0pt, leftmargin=10pt]
        \item \textit{KG knowledge utilization:} Compared to EHR-only methods (such as Direct Prompting), \modelname\ leverages structured KG knowledge to guide reasoning. In Case A, KG helps correct an incorrect prediction; in Case B, it assists in identifying a high-risk patient (\textcolor[HTML]{DAA520}{see words in yellow}).
        \item \textit{Hallucination reduction:} LLMs may hallucinate, but \modelname\ corrects these errors through KG-based verification. In Case A, a hallucinated statement in Prediction I is corrected in Stage II using KG-derived evidence (\textcolor[HTML]{228B22}{see words in green}).
    \end{itemize}

    \item \textbf{Augmented reasoning structure:} \modelname\ goes beyond baseline reasoning methods by incorporating structured and multi-stage reasoning:
    \begin{itemize}[topsep=0pt, itemsep=0pt, parsep=0pt, leftmargin=10pt]
        \item \textit{Structured reasoning:} \modelname\ mirrors human-like diagnostic reasoning. It summarizes patient information, retrieves relevant knowledge from the KG for inclusion criteria, prompts the LLM for prediction, and incorporates exclusion criteria to refine the decision.
        \item \textit{Multi-agent collaboration:} Unlike methods that merely extract KG paths as text prompts, \modelname\ employs a more sophisticated medical reasoning process. It utilizes a multi-agent framework that decomposes complex medical reasoning into specialized tasks, allowing collaboration between different agents. In both cases, the prompts integrate information from the KG, extracted by the linkage and retrieval agents, which collaborate with the prediction agent.
        \item \textit{Multi-stage reasoning:} \modelname\ adopts a multi-stage conversational reasoning approach to enhance prediction robustness. In both cases, the prediction follows a two-stage process, encouraging the LLM to consider different perspectives and refine its reasoning.
    \end{itemize}

    \item \textbf{Robust prediction performance:} \modelname\ generates accurate predictions and reasoning aligned with ground truth. In Case A, Stage I’s incorrect prediction is corrected in Stage II, while Case B’s initial prediction remains correct and robust.

\end{itemize}

\begin{tcolorbox}[colback=gray!10, colframe=black, title=Case A (Ground Truth as NO), width=\textwidth, fontupper=\scriptsize, boxsep=2pt, left=2pt, right=2pt, top=2pt, bottom=2pt]
    
    \begin{tcolorbox}[colback=red!10, colframe=black, title=Prediction Stage I Prompts, width=\textwidth-0mm, boxrule=0.2pt, fontupper=\scriptsize, boxsep=2pt, left=2pt, right=2pt, top=2pt, bottom=2pt]
        EHR Data: Nontraumatic subarachnoid hemorrhage; Other cerebrovascular diseases; Other functional intestinal disorders; acetaminophen-oxycodone; docusate; nimodipine; ondansetron; pravastatin.
        
        Question: Will the patient develop ``Post-stroke cognitive impairment'' at the next visit?
        
        Guidance from KG: Key diagnosis factors for PSCI include: Structural brain damage, neuroimaging evidence, and biochemical markers...
    \end{tcolorbox}

    \vspace{0pt} 

    \begin{tcolorbox}[colback=blue!10, colframe=black, title=Prediction I: YES, width=\textwidth-0mm, boxrule=0.2pt, fontupper=\scriptsize, boxsep=2pt, left=2pt, right=2pt, top=2pt, bottom=2pt]
        Reasoning: The patient has been diagnosed with nontraumatic subarachnoid hemorrhage and other cerebrovascular diseases, which are significant risk factors for PSCI ... Additionally, \textcolor[HTML]{228B22}{\textbf{pravastatin}} is linked to cognitive decline, further increasing the risk.
    \end{tcolorbox}

    \vspace{0pt}

    \begin{tcolorbox}[colback=red!10, colframe=black, title=Prediction Stage II Prompts, width=\textwidth-0mm, boxrule=0.2pt, fontupper=\scriptsize, boxsep=2pt, left=2pt, right=2pt, top=2pt, bottom=2pt]
        Instruction: Check your prediction cautiously. 
        
        Guidance from KG: Unrelated factors in diagnosing PSCI include a broad range of environmental toxins, pharmaceuticals, industrial chemicals, and biological compounds that do not contribute to cognitive dysfunction treatment or diagnosis. These factors include heavy metals (e.g., lead, cadmium, mercury), industrial pollutants (e.g., polychlorinated biphenyls, dioxins, benzene), pharmaceuticals (e.g., metformin, haloperidol, tacrine, \textcolor[HTML]{228B22}{\textbf{statins}})...
    \end{tcolorbox}

    \vspace{0pt}

    \begin{tcolorbox}[colback=blue!10, colframe=black, title=Prediction II: NO, width=\textwidth-0mm, boxrule=0.2pt, fontupper=\scriptsize, boxsep=2pt, left=2pt, right=2pt, top=2pt, bottom=2pt]
        Reasoning: While the patient has cerebrovascular diseases, the provided information does not confirm key diagnostic factors for PSCI. The diagnosis requires severe structural brain damage, biochemical markers, and neuroimaging evidence, none of which are explicitly mentioned. Additionally, \textcolor[HTML]{228B22}{\textbf{pravastatin}} and nimodipine do not directly contribute to PSCI risk. Given the standards provided, there is insufficient evidence to definitively predict PSCI.
    \end{tcolorbox}
    
\end{tcolorbox}

\vspace{5pt}
\begin{tcolorbox}[colback=gray!10, colframe=black, title=Case B (Ground Truth as YES), width=\textwidth, fontupper=\scriptsize, boxsep=2pt, left=2pt, right=2pt, top=2pt, bottom=2pt]
    
    \begin{tcolorbox}[colback=red!10, colframe=black, title=Prediction Stage I Prompts, width=\textwidth-0mm, boxrule=0.2pt, fontupper=\scriptsize, boxsep=2pt, left=2pt, right=2pt, top=2pt, bottom=2pt]
        EHR Data: Nephritis and nephropathy, not specified as acute or chronic, with lesion of proliferative \textcolor[HTML]{DAA520}{\textbf{glomerulonephritis}}; \textcolor[HTML]{DAA520}{\textbf{Angina}} decubitus; Diabetes mellitus without mention of complication, type II or unspecified type, not stated as uncontrolled; Background diabetic retinopathy; Cardiovascular Surgery Service...
        
        Question: Will the patient develop ``Chronic kidney disease'' at the next visit?
        
        Guidance from KG: Chronic kidney disease (CKD) is a progressive condition characterized by declining kidney function, often leading to kidney failure. It shares similarities with conditions such as membranous \textcolor[HTML]{DAA520}{\textbf{glomerulonephritis,}} hypertension, and focal segmental glomerulosclerosis. Key diagnostic indicators include ... Cardiovascular complications, including acute coronary syndrome, \textcolor[HTML]{DAA520}{\textbf{angina}}, heart murmurs, and high cardiac output, further indicate CKD progression. 
    \end{tcolorbox}

    \vspace{0pt} 

    \begin{tcolorbox}[colback=blue!10, colframe=black, title=Prediction I: YES, width=\textwidth-0mm, boxrule=0.2pt, fontupper=\scriptsize, boxsep=2pt, left=2pt, right=2pt, top=2pt, bottom=2pt]
        Reasoning: Nephritis and nephropathy with proliferative glomerulonephritis indicates kidney involvement and is a potential precursor to CKD if it progresses. Diabetes mellitus (Type II) is a major risk factor for CKD, as diabetic nephropathy is a leading cause of chronic kidney disease. Cardiovascular conditions (coronary atherosclerosis, angina, bypass surgery, PSVT) also indicate kidney disease since cardiovascular disease and CKD often coexist and contribute to each other’s progression. Diabetic microvascular complications (like retinopathy) are closely linked to kidney damage. Given the presence of glomerulonephritis, diabetes, and cardiovascular complications, the patient is at a high risk of developing CKD.
    \end{tcolorbox}

    \vspace{0pt}

    \begin{tcolorbox}[colback=red!10, colframe=black, title=Prediction Stage II Prompts, width=\textwidth-0mm, boxrule=0.2pt, fontupper=\scriptsize, boxsep=2pt, left=2pt, right=2pt, top=2pt, bottom=2pt]
        Instruction: Check your prediction cautiously. 
        
        Guidance from KG: Unrelated factors in diagnosing CKD include conditions such as hypertension ...
    \end{tcolorbox}

    \vspace{0pt}

    \begin{tcolorbox}[colback=blue!10, colframe=black, title=Prediction II: YES, width=\textwidth-0mm, boxrule=0.2pt, fontupper=\scriptsize, boxsep=2pt, left=2pt, right=2pt, top=2pt, bottom=2pt]
        Reasoning: The patient has nephritis and nephropathy, which indicate kidney damage, and proliferative glomerulonephritis, a condition that can lead to progressive kidney dysfunction. Additionally, diabetes mellitus and coronary atherosclerosis are well-known risk factors for CKD. Given these conditions, the patient is at a significant risk of developing chronic kidney disease.
    \end{tcolorbox}
    
\end{tcolorbox}

\section*{Conclusion}
In this work, we proposed \modelname, a KG-enhanced reasoning framework for zero-shot diagnosis prediction using LLMs. Our approach addresses key challenges in LLM-based medical diagnosis, including the need for structured reasoning, hallucination reduction, and improved effectiveness. By integrating a multi-agent framework consisting of a linkage agent, a retrieval agent, and a prediction agent, \modelname\ effectively connects patient data with structured medical knowledge and refines its diagnostic reasoning through multi-stage conversational steps. Experimental results demonstrate that \modelname\ enhances diagnostic accuracy compared to existing methods effectively and efficiently. Our findings highlight the potential of structured knowledge integration and iterative reasoning in medical AI, paving the way for more reliable and scalable zero-shot diagnosis prediction models. Future work may explore further optimizations in multi-agent collaboration and expanding KG-based reasoning to a broader range of medical conditions.


\makeatletter
\renewcommand{\@biblabel}[1]{\hfill #1.}
\makeatother

\bibliographystyle{vancouver}
\bibliography{reference}

\end{document}